\documentclass[journal]{IEEEtran}

\ifCLASSINFOpdf
\else
   \usepackage[dvips]{graphicx}
\fi
\usepackage{url}

\usepackage{hyperref}
\usepackage{graphicx}
\usepackage{amsmath}
\usepackage{float}
\usepackage{caption}
\usepackage{stfloats}
\usepackage{afterpage}
\usepackage{epstopdf}
\usepackage{booktabs}
\usepackage{array}
\usepackage{etoolbox}
\usepackage{algorithm}
\usepackage{algpseudocode}
\usepackage{arydshln}
\usepackage{booktabs}
\usepackage[table]{xcolor}

\usepackage{pgfplots}
\pgfplotsset{compat=1.17}

\usepackage{placeins}
\usepackage{tikz}
\usepackage{textcomp}
\usepackage{hyperref}
\usepackage{lipsum}

\begin{document}

\title{Simple Self-Organizing Map with Vision Transformers}

\author{Alan Luo, Kaiwen Yuan
\thanks{Alan Luo is with the Siebel School of Computing \& Data Sciences at the University of Illinois Urbana-Champaign, 201 N Goodwin Ave, Urbana, IL 61801, USA (e-mail: alanluo3@illinois.edu).}
\thanks{Kaiwen is with Safari AI Inc., 12 E 49th St. New York, NY 10017, USA (e-mail: kaiwen.yuan@getsafari.ai).}
}

\markboth{Journal of \LaTeX\ Class Files,~Vol.~18, No.~9, March~2025}
{Shell \MakeLowercase{\textit{et al.}}: Bare Demo of IEEEtran.cls for IEEE Journals}
\maketitle

\maketitle

\newcommand{\highlightbib}[1]{\color{red}#1\color{black}}

\begin{abstract}
Vision Transformers (ViTs) have demonstrated exceptional performance in various vision tasks. However, they tend to underperform on smaller datasets due to their inherent lack of inductive biases. Current approaches address this limitation implicitly—often by pairing ViTs with pretext tasks or by distilling knowledge from convolutional neural networks (CNNs) to strengthen the prior. In contrast, Self-Organizing Maps (SOMs), a widely adopted self-supervised framework, are inherently structured to preserve topology and spatial organization, making them a promising candidate to directly address the limitations of ViTs in limited or small training datasets. Despite this potential, equipping SOMs with modern deep learning architectures remains largely unexplored. In this study, we conduct a novel exploration on how Vision Transformers (ViTs) and Self-Organizing Maps (SOMs) can empower each other, aiming to bridge this critical research gap. Our findings demonstrate that these architectures can synergistically enhance each other, leading to significantly improved performance in both unsupervised and supervised tasks. Code is publicly available on GitHub\footnote{\url{https://github.com/aluo7/ViT-SOM}}.

\end{abstract}


\IEEEpeerreviewmaketitle

\section{Introduction}
The recent success of transformer architecture \cite{vaswani2017attention} in natural language processing (NLP) has quickly inspired the vision community to rethink vision tasks \cite{dosovitskiy2020vit, he2022mae} in a similar way. The Vision Transformer (ViT), introduced by \cite{dosovitskiy2020vit}, marks a significant shift in the paradigm of computer vision, transitioning from convolutional neural networks (CNNs) towards transformer-based architectures. Conventionally, ViTs model images as sequences of patches, analogous to the concept of tokens in NLP. This approach has demonstrated competitive performance on image classification tasks when compared to traditional CNNs, especially when pre-trained on large-scale datasets. However, a well-documented limitation of ViTs is their lack of inductive biases \cite{dosovitskiy2020vit}, resulting in poor performance when trained on limited or small training datasets. Various effective techniques \cite{liu2021efficient, das2024limited} have been proposed to address this challenge implicitly by learning strong priors, including methods such as pretext tasks \cite{liu2021efficient, das2024limited} or knowledge distillation \cite{li2022locality}. While these approaches are effective, their lack of inherent inductive biases presents an opportunity to explore new methods for directly embedding such biases into ViTs.

\begin{figure}[t]
\centerline{\includegraphics[width=\columnwidth]{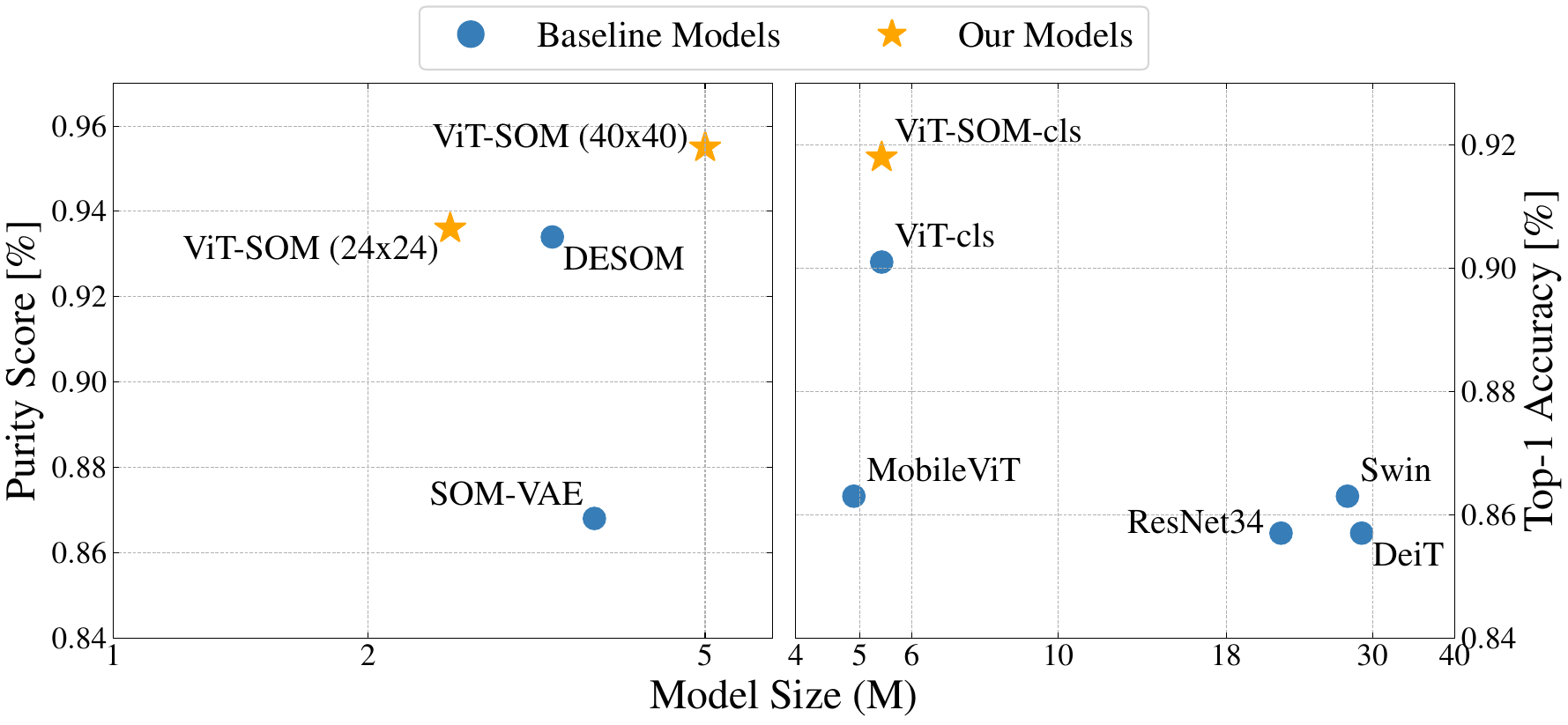}}
\caption{Model size (M) and task performances comparisons between ViT-SOM and baseline methods on MNIST and CIFAR-10 datasets. Model size is log scaled for interpretability.}
\end{figure}

On the other hand, self-organizing maps (SOMs) \cite{kohonen1982self}, a powerful self-supervised method, can intrinsically preserve topology and spatial organization as inductive biases, posing a natural solution to address the challenges posed by ViTs. However, while SOMs have been widely adopted in various data analysis applications \cite{wirth2011expression, wang2022implementing} over the past decades, classic SOMs suffer from a major limitation: poor feature abstraction capabilities. As such, ViTs serve as a strong complement to SOMs, addressing their weakness in feature abstraction while benefiting from SOMs inherent inductive biases.

While this hypothesis is promising, prevailing research has yet to explore this direction. The only related work we are aware of is TENSO \cite{yoon2023anomalous}, which focuses on low-dimensional trajectory data. The majority of relevant publications have focused on CNN based architectures or variants \cite{lichodzijewski2002host, liu2015deep, ferles2018denoising, fortuin2019somvae, forest2019deep, forest2021desom} to enhance SOMs feature extraction. For example, one pioneering work, DSOM \cite{liu2015deep}, demonstrated a significant 7.17\% performance improvement over classic SOMs on the MNIST dataset. More recent works have focused on handling sequential data with Long Short-Term Memory (LSTM) \cite{forest2019deep, manduchi2021t} and Contrastive Predictive Coding (CPC) \cite{huijben2023som}, rather than leveraging more powerful architectures like ViTs.

Therefore, there is a clear research gap in studying the underexplored, mutually beneficial interactions between SOMs and ViTs. In our work, we propose ViT-SOM, a novel framework that integrates ViTs with SOMs to leverage the strengths of both architectures. Our approach involves two steps: 1) evaluating the unsupervised setting, similar to \cite{fortuin2019somvae, forest2021desom}, for clustering tasks on various datasets; 2) extending the experiments to supervised classification on small datasets. As detailed in later sections, our simple yet effective method achieves leading performance in both settings on various benchmark datasets. We believe our work can serve as a strong foundation to initiate this innovative research direction.

\section{Proposed Method}
\label{proposed_method}

\begin{figure*}[t]
\centerline{\includegraphics[width=\linewidth]{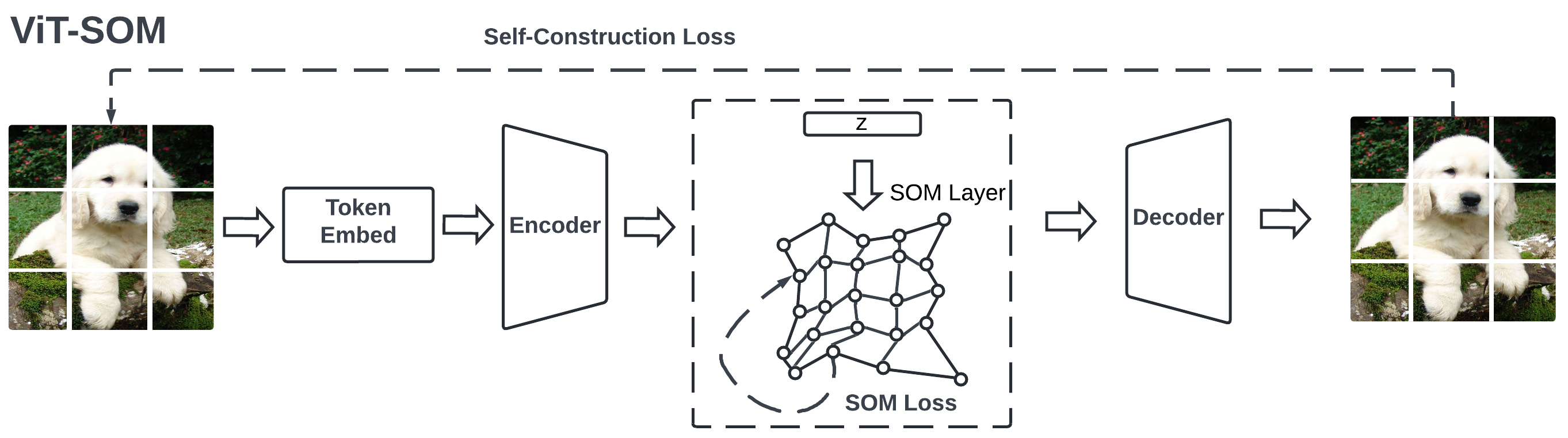}}
\caption{ViT-SOM architecture.}
\label{fig:arch}
\end{figure*}

\subsection{SOM Debrief}
SOMs utilize a competitive learning strategy, where each unit—often referred to as a prototype or neuron—competes to represent the input data. For a given input $z$ and a weight matrix $W_{som}$, the best matching unit (BMU) is identified at each cycle based on spatial distance. Once the BMU is determined, weights are updated according to the neighborhood (update) function, which considers both spatial distance per sample and learning rate (with decay). A naive SOM update process from step $k$ to $k+1$ can be described as follows:
\begin{equation} \label{eq:som_update}
W_{som, k+1} = W_{som, k} + \alpha_{k} \cdot h_{k, i, j} (z, W_{som, k})
\end{equation}
where $\alpha_{k}$ is the learning rate and $h_{k, i, j}$ denotes the neighborhood function at step $k$ between the BMU $j$ and an arbitrary neuron $i$. In practice \cite{forest2021desom}, a common approach to the neighborhood function \( h_{k, a, b} \) is to employ a Gaussian neighborhood function with a distance metric between two neurons:
\begin{equation} \label{eq:n_func}
h_{k, i, j} = \exp\left(-\frac{\sum_{i=1}^n d_{ij}}{2T(k)^2}\right)
\end{equation}
where $T(k)$, often termed temperature, decays with time, and $d_{ij}$ represents the distance metric between the BMU and an arbitrary neuron.

For the SOM layer, we adopt a standard self-organizing map governed by Equations \ref{eq:n_func} and \ref{eq:loss_som}. The influence of the neighborhood is controlled by a temperature parameter $T(k)$, which decays exponentially over $K$ iterations as depicted below:

\begin{equation}
T(k) = T_{\text{max}} \left( \frac{T_{\text{min}}}{T_{\text{max}}} \right)^{\frac{k}{K}}
\label{eq:temp_decay}
\end{equation}

where $k$ is the current training iteration. Following DESOM \cite{forest2021desom}, we employ an exponential temperature decay. Initial temperature $T_{\text{max}}$ is set to be half the map size, a well-explored best practice for larger maps \cite{tallec2006self} that encourages broader initial prototype separation and improved stability of the final quantization. Final temperature is set to a small constant $T_{\text{min}}=0.001$ as explored in Table II of our supplementary material, enabling fine-tuned updates to prototypes in late-stage training.

\subsection{ViT-SOM}

Classical SOMs update prototypes sequentially \cite{cottrell1998theoretical}, which is computationally inefficient and incompatible with GPU parallelization. To address this, we adopt a batch-compatible framework \cite{forest2021desom} where BMUs for all samples are computed in parallel, addressing SOMs inherently autoregressive training updates. Prototypes are then optimized via backpropagation on the loss:
\begin{equation} \label{eq:loss_som}
L_{som} = \frac{1}{N} \sum_{i=1}^{N} \sum_{j=1}^{M} w_{ij} \cdot d_{ij}(z)
\end{equation}
where $z$ is the input data point (in our context, an embedding space element), \( N \) is the number of input samples, \( M \) is the number of SOM units, \( w_{ij} \) represents the weight based on the distance to the BMU, and \( d_{ij} \) is the distance between input sample \( i \) and SOM unit \( j \). Despite their uncontested use in typical SOM architectures \cite{fortuin2019somvae, forest2021desom}, both the Euclidean and Manhattan distance functions suffer largely from scale variance. To address this limitation, we instead apply cosine similarity for $d_{ij}$:
\begin{equation} \label{eq:d_fcn}
d_{ij}(z) = 1 - \frac{z \cdot w_{ij}}{\|z\| \|w_{ij}\|}
\end{equation}

Our choice to utilize cosine similarity is motivated by the curse of dimensionality, a well-explored challenge when working with high-dimensional embeddings. Euclidean and Manhattan distances become less informative as dimensionality increases—a weakness imposed by ViTs high-dimensional latent space. We empirically validate this design choice in Table I of our ablation studies (see supplementary material), demonstrating that Cosine Similarity provides a more stable and meaningful signal for the SOM to constrain the latent space.

Figure \ref{fig:arch} depicts the architecture of ViT-SOM. The major difference lies in the embedding layer. Instead of naively passing the embedding vector $z$ to the decoder, ViT-SOM introduces an SOM layer to self-supervise the embedding vector for topology-preserving training:
\begin{equation}
J = \sum_{z \in \mathcal{Z}} \min_{i,j,N} (1 - \frac{z \cdot w_{ij}}{\|z\| \|w_{ij}\|})
\end{equation}
Since the SOM has a grid topology, this formulation ensures that the embedding vector $z$ is spatially projected onto the BMUs, preserving the underlying structure of the data throughout training. The iterative neighborhood updates enforce topological consistency across the latent space, ultimately leading to enhanced feature representation and improved data organization.

We utilize a tiny version of the Vision Transformer (ViT) \cite{touvron2021deit}, with detailed configurations for clustering and classification provided in Table \ref{tab:vit_configs}. To clarify our methodology, we present the high-level architecture of ViT-SOM in algorithm \ref{alg:vit_som_pseudocode}.

In both supervised and unsupervised settings, the objective functions are weighted by $\gamma$, which functions as a tunable hyperparameter to balance loss signals.

\begin{equation} \label{eq:loss}
L_{total} = L_{nn} + \gamma \cdot L_{som}
\end{equation}

\noindent $L_{nn}$ is the loss of the deep neural network. We set $\gamma=0.005$ for the clustering case and $\gamma=0.01$ for classification as explored in Table I of our ablation studies (see supplementary material), and further employ a linear warmup on lambda to prioritize feature learning at early stages over topological organization.

\begin{figure*}
    \centering
    \includegraphics[scale=0.175]{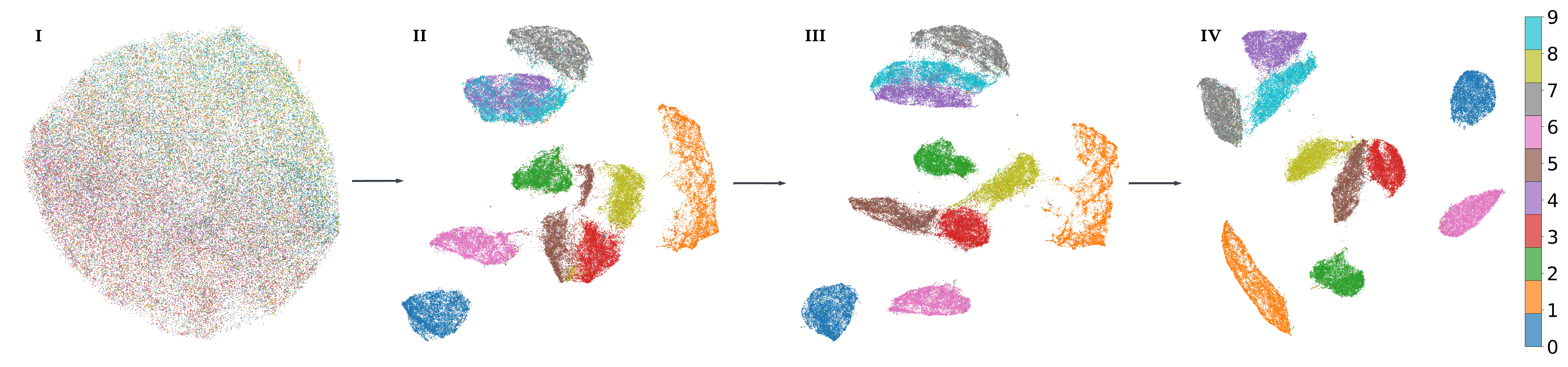}
    \caption{UMAP visualization of latent space embeddings from ViT-SOM model training progression on CIFAR-10. Projections shown at epochs 0, 25, 50, 200. Color mapping represents ground truth classes, demonstrating emergent cluster separation during training.}
\label{fig:umap}
\end{figure*}

\begin{table}[t]
    \centering
    \caption{ViT Backbone configurations for ViT-SOM on classification and clustering tasks.}
    \label{tab:vit_configs}
    \begin{tabular}{lccccc}
        \toprule
        \textbf{Task} & \textbf{Embed.} & \textbf{MLP} & \textbf{Enc./Dec.} & \textbf{Heads} & \textbf{Params} \\
        & \textbf{Dim.} & \textbf{Dim.} & \textbf{Depth} & & \\
        \midrule
        Classification & 192 & 768 & 12 / 2 & 3 & 5.4M \\
        Clustering & 16 & 64 & 4 / 2 & 2 & 2.5M \\
        \bottomrule
    \end{tabular}
\end{table}

\begin{algorithm}[h!]
\caption{ViT-SOM Pseudocode}
\label{alg:vit_som_pseudocode}
\begin{algorithmic}[1]
\footnotesize
\State **Initialize:** $ViT$ (Encoder \& Decoder), $SOM$ (with prototypes $P$)
\Procedure{Forward}{$x$}
    \State $z_{cls}, z_{patches}, x_{recon} \leftarrow ViT(x)$
    \State $z_{som} \leftarrow \text{flatten}(z_{patches})$
    \State $distances, bmu\_indices \leftarrow SOM(z_{som})$
    \If{classification task}
        \State $L_{nn} \leftarrow \text{ClsLoss}(\text{ClsHead}(z_{cls}), y_{true})$
    \Else
        \State $L_{nn} \leftarrow \text{ReconLoss}(x_{recon}, x)$
    \EndIf
    \State $L_{SOM} \leftarrow \text{SOMLoss}(distances, bmu\_indices)$
    \State $L_{total} \leftarrow L_{nn} + \gamma \cdot L_{SOM}$
\EndProcedure
\end{algorithmic}
\end{algorithm}

\section{Experiments and Results}
\subsection{Training}

The pipeline is implemented in Pytorch and a single NVIDIA RTX 4090 GPU is used for training. AdamW \cite{loshchilov2018adamw} is employed as the optimizer with an initial learning rate of $lr=0.01$, decayed via cosine annealing, and hyperparameters \( \beta_1 = 0.9 \), \( \beta_2 = 0.999 \) for all configurations. For classification tasks, we simply adopt the MAE \cite{he2022mae} data augmentation configuration. For clustering tasks, no data augmentation is used (e.g., random cropping, flipping, and color jitter). All models are trained from scratch, and no pre-trained weights are used during initialization. A notable drawback of ViT-SOM is the trade-off in inference latency. However, considering the small number of trainable parameters, we believe this latency can be optimized. Future work could focus on more efficient implementations of the BMU search, such as exploring more parallelizable algorithms. Despite this latency, a thorough computational benchmark demonstrates that the model remains highly competitive across other key metrics, including training time and GPU memory usage. Results are provided in the supplementary material.

\subsection{Datasets.}

\paragraph{Unsupervised} 
We evaluate our unsupervised learning experiments on the MNIST \cite{lecun1998mnist}, Fashion-MNIST \cite{xiao2017fashionmnist}, and USPS \cite{hull1994usps} datasets so that we can fairly compare with DESOMs settings \cite{forest2021desom}. The MNIST and Fashion-MNIST datasets consists of 70,000 $28 \times 28$ grayscale images of handwritten digits (0-9) and fashion items across 10 categories respectively, split into 60,000 training images and 10,000 testing images. The USPS dataset contains 9,298 $16 \times 16$ grayscale images of handwritten digits (0-9).

\paragraph{Supervised} For classification tasks, we evaluate model performance on small natural image datasets, including CIFAR-10, CIFAR-100  \cite{krizhevsky2009cifar}, Flowers17 \cite{Nilsback06}, SVHN \cite{netzer2011svhn}, Tiny ImageNet \cite{Le2015TinyIV}, and MedMNIST \cite{medmnistv1}. Both CIFAR-10 and CIFAR-100 contain 60,000 $32 \times 32$ color images, with CIFAR-10 spanning 10 classes and CIFAR-100 containing 100 classes. Flowers17 comprises 1,360 color images of flowers across 17 categories (80 images per class). The SVHN dataset provides $32 \times 32$ color images of house numbers from street view imagery, organized into 10 classes. Tiny ImageNet \cite{Le2015TinyIV} is a subset of the well-known ImageNet-1k with 100,000 images and 200 classes. MedMNIST \cite{medmnistv1} consists of a collection of standardized biomedical images, offering a demonstration of ViT-SOM in a practical setting with 107,180 images and 9 classes.

\subsection{Unsupervised Clustering Results}

The widely recognized purity score metric is adopted to evaluate the clustering performance across datasets. Table \ref{tab:clustering} presents clustering benchmark results for various SOM-equipped models, including SOM-VAE \cite{fortuin2019somvae}, DESOM \cite{forest2021desom}, and our proposed ViT-SOM with varying map sizes. ViT-SOM achieves significantly higher purity scores on MNIST and Fashion-MNIST compared to SOM-VAE while using fewer learnable parameters. Notably, ViT-SOM (24$\times$24) outperforms DESOM—a CNN-based variant—across all datasets with 24\% fewer parameters. Additionally, ViT-SOM (40$\times$40) outperforms DESOM by an average 14.2\% in purity scores, demonstrating ViTs’ strength as a feature extractor in SOM frameworks.

\begin{table*}[t]
    \centering
    \caption{SOM-Enhanced ViT Clustering Performance Comparisons.}
    \label{tab:clustering}
    \begin{tabular}{lcccc}
        \toprule
        \textbf{Model} & \textbf{MNIST} & \textbf{Fashion-MNIST} & \textbf{USPS} & \textbf{\# Params} \\
        \midrule
        SOM (24$\times$24) & 0.711$\pm$0.005 & 0.668$\pm$0.002 & 0.732$\pm$0.009 & 451K \\
        SOM-VAE \cite{fortuin2019somvae} & 0.868$\pm$0.03 & 0.739$\pm$0.002 & 0.854$\pm$0.010 & 3.7M \\
        DESOM \cite{forest2021desom} & 0.934$\pm$0.004 & 0.751$\pm$0.009 & 0.857$\pm$0.011 & 3.3M \\
        ViT-SOM (24$\times$24) & 0.936$\pm$0.003 & 0.817$\pm$0.002 & 0.935$\pm$0.006 & 2.5M \\ 
        ViT-SOM (40$\times$40) & \textbf{0.955$\pm$0.001} & \textbf{0.841$\pm$0.006} & \textbf{0.948$\pm$0.025} & 5.0M \\ 
        \bottomrule
    \end{tabular}
\end{table*}

\begin{table*}[t]
    \centering
    \caption{SOM-Enhanced ViT Classification Performance Comparisons.}
    \label{tab:classification}
    \begin{tabular}{lccccccc}
        \toprule
        \textbf{Model} & \textbf{CIFAR-10} & \textbf{CIFAR-100} & \textbf{Flowers17} & \textbf{SVHN} & \textbf{Tiny-ImageNet} & \textbf{MedMNIST} & \textbf{\# Params} \\
        \midrule
        ResNet34 \cite{Chen2023} & 0.857$\pm$0.005 & 0.638$\pm$0.004 & 0.783$\pm$0.011 & 0.894$\pm$0.004 & 0.468$\pm$0.011 & 0.826$\pm$0.008 & 21.8M \\
        Swin-T \cite{liu2021swin} & 0.863$\pm$0.003 & 0.597$\pm$0.003 & 0.886$\pm$0.010 & 0.952 $\pm$0.001 & 0.501$\pm$0.014 & 0.842$\pm$0.005 & 27.5M \\
        DeiT-T \cite{touvron2021deit} & 0.857$\pm$0.009 & 0.608$\pm$0.003 & 0.882 $\pm$0.013 & 0.958$\pm$0.007 & 0.499$\pm$0.009 & 0.838$\pm$0.007 & 5.7M \\
        MobileViT-T \cite{mehta2022mobilevit} & 0.863$\pm$0.010 & 0.681$\pm$0.007 & 0.739 $\pm$0.071 & 0.946$\pm$0.003& 0.452$\pm$0.018 & 0.840$\pm$0.006 & 4.9M \\
        ViT-cls (reproduce) & 0.901$\pm$0.001 & 0.604$\pm$0.009 & 0.884$\pm$0.006 & 0.957$\pm$0.006 & 0.496$\pm$0.005 & 0.835$\pm$0.006 & 5.4M \\
        ViT-SOM-cls & \textbf{0.920$\pm$0.003} & \textbf{0.683$\pm$0.006} & \textbf{0.917$\pm$0.002} & \textbf{0.966$\pm$0.001} & \textbf{0.502$\pm$0.008} & \textbf{0.844$\pm$0.004} & 5.4M \\
        \bottomrule
    \end{tabular}
\end{table*}

Figure \ref{fig:umap} visualizes the evolution of the ViT-encoded latent space via UMAP projections throughout training. As training progresses, the ViT-SOM objective drives embeddings to form well-separated clusters, with points colored by digit class coalescing into distinct groups. This demonstrates the capability of our proposed method to organize latent representations semantically. Here, semantically similar digits—such as ($0$ and $6$), ($3$, $5$, and $8$), or ($4$, $7$, and $9$)—are topologically grouped as neighbors, directly reflecting the organizing influence of the SOM objective.

While the SOM objective successfully organizes the latent space, limitations inherent to the fixed grid structure persist. These artifacts (e.g., ambiguous $8$s surrounded by $5$s and $3$s), arising from the rigidity and sensitivity to initialization, can occasionally trap prototypes in suboptimal regions of the latent space. Future work could explore adaptive grid topologies to refine these boundary regions, further enhancing the organization demonstrated in the UMAP visualization.

\subsection{Classification Results}

The ViT-SOM-based classifier (denoted as ViT-SOM-cls) is evaluated against various established baseline models, with key results summarized in Table \ref{tab:classification}.

Our experiments demonstrate three key advantages of the proposed framework. First, ViT-SOM-cls achieves state-of-the-art accuracy when trained from scratch across all datasets. Notably, ViT-SOM-cls demonstrates superior efficiency, outperforming contemporary ViT variants and larger CNN-based models across all datasets. For instance, it outperforms the Swin Transformer \cite{liu2021swin} by over 14\% on CIFAR-100 and ResNet34 \cite{Chen2023} by over 17\% on Flowers17. This performance is achieved while requiring up to 79\% fewer trainable parameters than other architectures on average. Finally, when compared to ViT-cls, our reproduced baseline, ViT-SOM-cls exhibits improvements across the board, demonstrating the clear advantages of introducing a SOM as an inductive bias.

\section{Conclusion}

This work addresses two critical limitations in modern deep learning: Vision Transformers' (ViTs) lack of inherent inductive biases for limited or small datasets, and Self-Organizing Maps' (SOMs) suboptimal feature abstraction capabilities. We bridge this gap by proposing ViT-SOM, a framework that synergistically integrates ViTs and SOMs to mutually enhance their strengths. ViT-SOM leverages SOMs to impose topological constraints on ViT embeddings, while ViTs empower SOMs with robust feature extraction in high-dimensional spaces. Empirically validated in both supervised and unsupervised settings, our method demonstrates consistent improvements over baseline and state-of-the-art models, including computationally heavier architectures like DeiT, Swin, and ResNet34. Crucially, ViT-SOM achieves this without complex architectural modifications, highlighting the inherent compatibility of these frameworks. By demonstrating the mutual enhancement of ViTs and SOMs, this work encourages broader research in this area.\newline

\bibliographystyle{IEEEtran.bst}
\bibliography{biblio}

\end{document}